%% file: full-paper.tex
\documentclass{book}    

\usepackage{piers}  
\usepackage[]{graphicx}
\graphicspath{{Figures/}}
\usepackage{graphicx}

\usepackage{amsmath}
\usepackage{amssymb}
\usepackage{epsfig}
\usepackage{cite}
\usepackage{color}
\usepackage{balance}
\usepackage{pgfplots}
\usepackage{wrapfig}
\usepackage{lipsum}
\usepackage{hyperref}
\usepackage{gensymb}
\usepackage{bm}
\usepackage{breqn}
\usepackage{enumitem}
\usepackage{tabto}

\begin{document}

\title{Interpreting a Semantic Segmentation Model for Coastline Detection }
\maketitle

\author      {F. M. Lastname}
\affiliation {University}
\address     {}
\city        {Boston}
\postalcode  {}
\country     {USA}
\phone       {345566}    
\fax         {233445}    
\email       {email@email.com}  
\misc        { }  
\nomakeauthor

\author      {F. M. Lastname}
\affiliation {University}
\address     {}
\city        {Boston}
\postalcode  {}
\country     {USA}
\phone       {345566}    
\fax         {233445}    
\email       {email@email.com}  
\misc        { }  
\nomakeauthor

\begin{authors}

{\bf Conor~O'Sullivan}$^{1,2}$ $^{\dagger}$, 
{\bf Seamus~Coveney}$^{3}$, 
{\bf Xavier~Monteys}$^{4}$, 
{\bf and Soumyabrata Dev}$^{1,2}$\\
\medskip

$^{1}$ADAPT SFI Research Centre, Dublin, Ireland\\

$^{2}$School of Computer Science, University College Dublin, Ireland \\

$^{3}$Envo-Geo Environmental Geoinformatics, Skibbereen, Ireland \\

$^{4}$Geological Survey Ireland, Dublin, Ireland

$^{\dagger}$ Presenting author and corresponding author

\end{authors}

\begin{paper}

\begin{piersabstract}
We interpret a deep-learning semantic segmentation model used to classify coastline satellite images into land and water. This is to build trust in the model and gain new insight into the process of coastal water body extraction. Specifically, we seek to understand which spectral bands are important for predicting segmentation masks. This is done using a permutation importance approach. Results show that the NIR is the most important spectral band. Permuting this band lead to a decrease in accuracy of 38.12 percentage points. This is followed by Water Vapour, SWIR 1, and Blue bands with 2.58,  0.78 and 0.19 respectively. Water Vapour is not typically used in water indices and these results suggest it may be useful for water body extraction. Permuting, the Coastal Aerosol, Green, Red, RE1, RE2, RE3, RE4, and SWIR 2 bands did not decrease accuracy. This suggests they could be excluded from future model builds reducing complexity and computational requirements.  

\end{piersabstract}

\psection{Introduction}
Interpretability is an important aspect of machine learning in remote sensing. It means that researchers and industry professionals can understand the reasoning behind model predictions. In remote sensing, the data is often complex and underlying processes are not fully understood. By understanding the reasons for predictions, we can gain new insight into the process being studied. Researchers and professionals are also more likely to trust the results of a model if we provide reasons for them. Finally, interpretability can help identify reasons for errors in predictions and help improve the accuracy of future models.

A particular question we seek to answer is which spectral bands are most important to the predictions of a deep learning model. The model is used for the semantic segmentation of coastline images--- classifying each pixel into land or water. By understanding this, we can relate the results to existing research on spectral bands for water body extraction. The hope is that this will provide valuable insight for future model builds. 

\psection{Related Work}

\input{figures/indices}

Previous research has focused on quantifying the individual predictive capability of spectral bands or indices using statistical methods. In terms of individual bands, the NIR band has shown to be effective for water extraction~\cite{mondejar2019near}.  Similarly, indices developed for water body extraction have been listed in Table~\ref{tab:indices}. In particular, SWI was developed for automated coastline detection. These indices are calculated by combining multiple spectral bands. When interpreting a deep learning model used for water extraction, we can expect similar bands to be important.

When training a deep learning model, it is not necessary to calculate indices beforehand. This is because all available spectral bands can be used as input into the model ~\cite{seale2022swed}. Through the process of training, the most important bands and relationships between the bands are used to make predictions. In some sense, the most appropriate indices can be encoded in the model's architecture. However, due to the black-box nature of deep learning models, these relationships can be difficult to uncover. 

There is research on interpreting deep learning models for semantic segmentation. Researchers have considered parameter gradients~\cite{vinogradova2020towards} and SHAP values~\cite{dardouillet2022explainability}. These approaches aim to understand which individual pixels are most important for segmentation predictions. Additionally, they do not use satellite images as input. We want to understand which spectral bands are important. This involves understanding the overall importance of the band and not the individual pixels in the bands. 

\psection{Contributions}
 Our main contributions are to:
\vspace{-0.2cm}
\begin{itemize}
    \item Replicate a deep learning model in previous work used for coastal water body extraction~\cite{seale2022swed}.  
    \vspace{-0.2cm}
    \item Interpret the model using a permutation feature importance approach to understand which spectral bands are most important to predictions. 
    \vspace{-0.2cm}
    \item We open source the code and model to reproduce the results of this paper. These can be found in the GitHub repository~\footnote{In the spirit of reproducible research, the codes and model to reproduce the results of this paper can be found here: \url{https://github.com/conorosully/interpreting-coastline-unet}.}
\end{itemize}

\psection{Methodology}

\psubsection{Dataset}
We base our analysis on the Sentinel-2 Water Edges Dataset (SWED)~\cite{seale2022swed}. This contains Sentinel-2 satellite images and binary segmentation masks from a variety of coastlines. This includes 28,224 training observations. Each observation includes 12 spectral bands as input and a binary segmentation mask, where each pixel was classified as land (0) or water (1), as output. These were annotated using a semi-supervised approach which used Coastal Aersoal, green, red, NIR and SWIR1 bands as input into a clustering algorithm. The dataset also contains 98 test images from 49 testing locations. These were manually annotated. 

\psubsection{Model Architecture and Training}
As seen in Figure~\ref{fig:unet}, an adaption of the U-Net architecture has been used~\cite{ronneberger2015u}. The architecture consists of an encoder, bottleneck and decoder. The encoder has four convolutional blocks (conv). Each block contains two convolutional layers with a 3 x 3 kernel followed by an Exponential Linear Unit (ELU) activation and then a batch normalisation layer. Finally, each block is down-sampled by a 2 x 2 max-pooling layer. The bottleneck has the same structure except for the max-pooling layer. The decoder has four blocks (deconv). These are symmetrical to the encoder blocks. Skip connections concatenate the final convolutional layer in each convolution block to the first layer in the decoder blocks. This first layer is a 2x2 deconvolutional layer followed by two convolutional layers with a 3 x3 kernel, ELU activation and batch normalisation. The last encoder block ends with a softmax activation function. 

This architecture was selected because we wanted to replicate the model architecture and training process in previous research ~\cite{seale2022swed}. This research experimented with different cost functions. We used the one that provided the best results based on the metrics described in the Evaluation Metrics section. This is the cross entropy loss function.  We use a training and validation set of sizes 25,401 and 2,823 respectively. The model was trained for 50 epochs and we select the one that had the lowest validation loss.

\begin{figure}[h]
\centering
\includegraphics[width=0.99\textwidth]{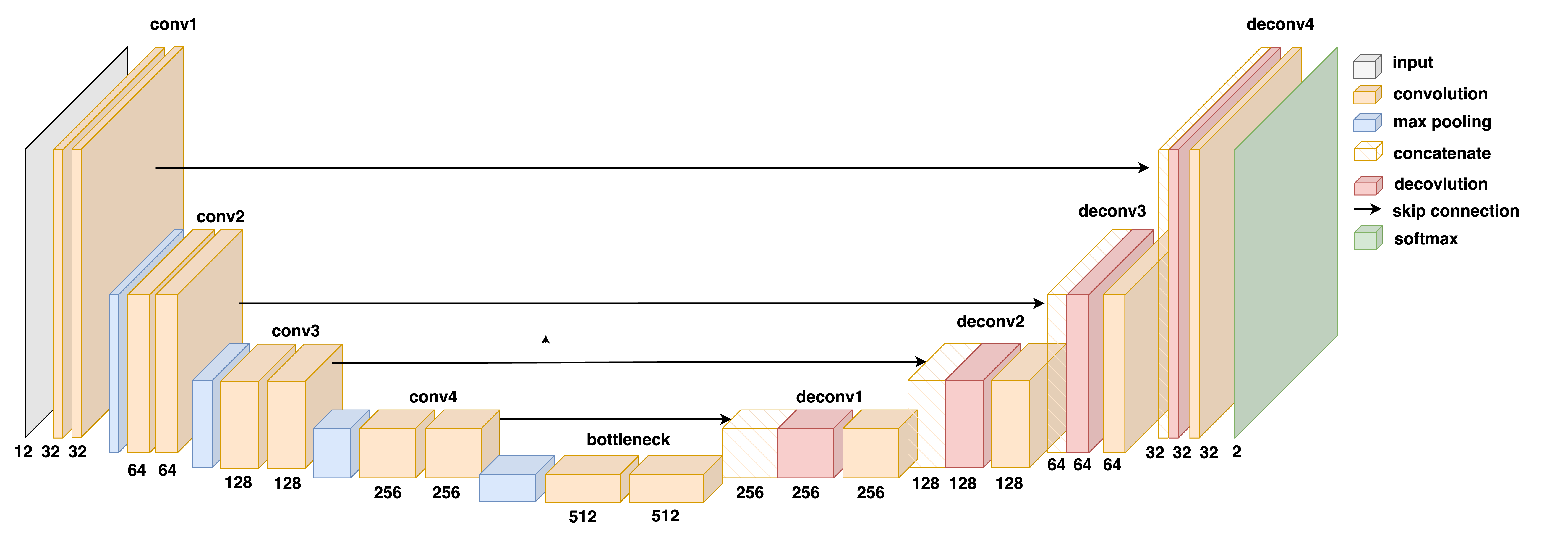}
\caption{Model architecture consisting of an encoder, bottleneck and decoder. Skip connections are used to concatenate layers in the encoder to symmetrical layers in the decoder. The number below each layer is the number of channels in that layer.}
\label{fig:unet}
\end{figure}

There were some aspects of the training process that we were not able to replicate exactly. This includes the training batch size used and the exact way the input has been scaled. Hence, we decided on a batch size of 32 and we scaled the input by dividing each pixel value by 10,000 and reduced any pixels greater than 1 to 1. There were also differences in the training and validation set sizes. Previous research had a smaller size of 23,807 and 2,661 respectively. It is not clear which training inputs were removed from SWED. 

\psubsection{Evaluation Metrics}

\input{figures/confusion_matrix}

To confirm that our model produces similar results, we evaluate it with metrics based on the confusion matrix in~\ref{tab:confusion_matrix}. Water is taken as the positive class and land as the negative class. Hence, TP and FP are the numbers of correctly and incorrectly labelled water pixels. TN and FN are the numbers of correctly and incorrectly labelled land pixels. For each of the 98 test images, we calculate the metrics listed below and take the average for each metric as the final value. 

 $$Accuracy = \frac{TP+TN}{TP+TN+FP+FN}$$
$$Balanced Accuracy = 0.5 \left (\frac{TP}{TP+FN} + \frac{TN}{TN+FP} \right )$$
$$      Precision = \frac{TP}{TP+FP} $$
$$       Recall = \frac{TP}{TP+FN} $$
$$        F1 = 2 \left(\frac{Precision*Recall}{Precision+Recall} \right) $$

\psubsection{Permutation importance}
We interpret this model using a permutation importance approach. We identified 3 images, listed in the appendix, with erroneous annotations. To give us a reliable comparison to previous work, we removed these from the test set leaving us with 95 images. These three images were not removed from the above validation step. Hence, permutation importance scores are calculated for a spectral band or combinations of bands by:
\begin{enumerate}
    \item Permuting the respective band(s) in each of the 95 test images. This is done by randomly shuffling the pixels within an image's band. 
    \item Using the permuted image as input, the trained U-Net is used to predict a binary segmentation mask. 
    \item Calculate the average accuracy over the 95 images. 
    \item Calculate the decrease in average accuracy when compared to the average accuracy obtained using the original 95 images.
\end{enumerate}
We repeated the above process 5 times and take the average decrease in accuracy as the final importance score. We calculate these permutation scores for each of the 12 spectral bands. We also include combinations of bands that correspond to the water extraction indices mentioned in Table~\ref{tab:indices} and the three visible light bands. 

\psection{Results and Discussion}
\psubsection{Evaluation Metrics}

The evaluation results are presented in Table~\ref{tab:eval}. We note differences between the original and the replicated model results. We consider these differences to be insignificant and are likely due to the aspect of the modelling development process that could not be replicated exactly. We, therefore, expect the interpretation of the replicated model to be similar to an interpretation of the original model. 

\input{figures/eval_results}

\psubsection{Permutation Importance of Individual Bands}

The permutation importance of each spectral band is given in Figure~\ref{fig:perm}.  These are calculated by permuting only the given band. The three largest importance scores are given for NIR, Water Vapour and SWIR 1. Permuting these bands lead to a decrease in accuracy of 38.12, 2.58 and 0.78 percentage points. The remaining bands provided a small ($<0.5$) positive or negative percentage point change. 

\begin{figure}[h]
\centering
\includegraphics[width=0.99\textwidth]{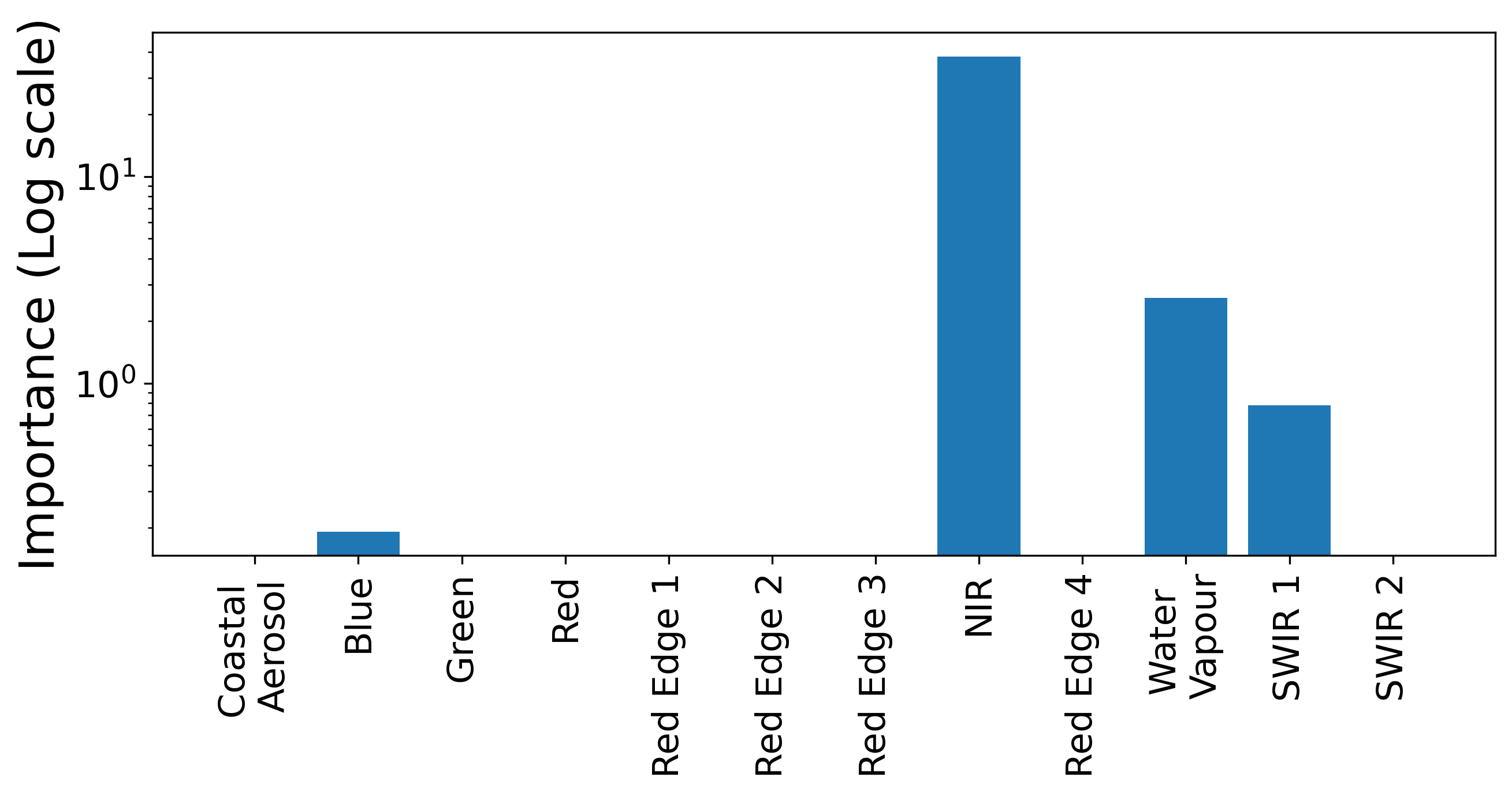}
\caption{Permutation importance scores for spectral bands used as input into the coastal segmentation model. The scores give a decrease in average accuracy when the respective band is permuted.}
\label{fig:perm}
\end{figure}

Based on these importance scores, NIR is the most important band for predicting the segmentation masks. Its score was over 14 times higher than the second-highest score for Water Vapour. We should consider that this may be a result of NIR being one of the bands used in the semi-supervised process for labelling the training set. In other words, the importance may reflect a bias towards this band in the training set and not the intrinsic predictive capability of this band. Similarly, SWIR 1 was also used in the semi-supervised process.

On the other hand, Water Vapour was not used in this process. Yet, based on this analysis it is the second most important spectral band. This result is particularly interesting in light of previous research. In Table~\ref{tab:indices}, we saw that it was not typical to use this band in indices for water extraction. This suggests that Water Vapour may have been overlooked when developing  these indices. 

\psubsection{Permutation Importance of Indices}

One consideration when analysing the above results is that spectral bands can provide similar information. In other words, the pixel values in the bands can be correlated. A robust model may use the information from all bands even if they provide similar information. As a result, permuting one band may not lead to a significant decrease in accuracy. This is because the model can still use the information in the remaining bands to make predictions. 

One way to test this hypothesis is by permuting multiple bands simultaneously. For example, the "Visible Light" bar in Figure~\ref{fig:perm_index} gives the importance score when we permute the Blue, Green and Red spectral bands. This leads to a decrease in accuracy of 0.25 percentage points. When we permutated these bands individually the accuracy changed by 0.19, -0.1 and  -0.1 percentage points respectively. Based on this, neither of these bands is particularly important to the prediction. Additionally, when combined their importance is not significantly improved. 

Similarly, the "Not Important" band gives the importance when we permute all the bands that did not have a positive importance in Figure~\ref{fig:perm}. This includes Coastal Aerosol, Green, Red, RE1, RE2, RE3, RE4 and SWIR 2. When permuted together, these bands changed the average accuracy by -0.47 percentage points. Based on this, we can be more certain that the model does not rely on these bands to make predictions.

\begin{figure}[h]
\centering
\includegraphics[width=0.99\textwidth]{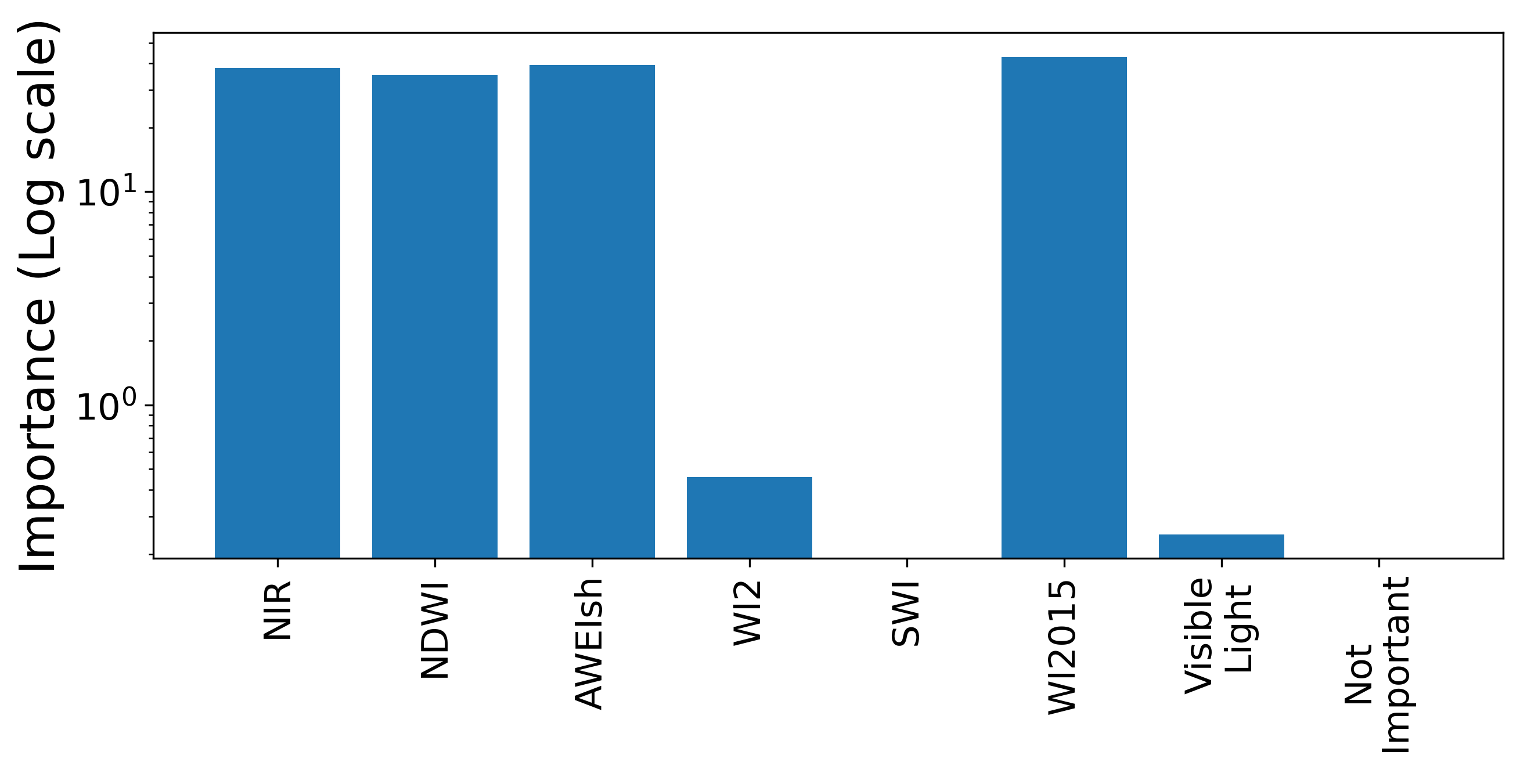}
\caption{Permutation importance scores for coastline segmentation model. The scores give the percentage change in average accuracy when combinations are bands are permuted.}
\label{fig:perm_index}
\end{figure}

The remaining bars in Figure~\ref{fig:perm_index} give the importance score if we permute the bands in the indices listed in Table~\ref{tab:indices}. This provides evidence for the effectiveness of these indices for water body extraction. Although we should keep in mind that this is not an indication of the individual predictive capability of the indices. It is only an indication that the model has used the bands in the given index's calculation when making predictions. For example, SWI is calculated using RE1 and SWIR2. In Figure~\ref{fig:perm} we saw that neither of these bands was important when permuted individually. Similarly, the NDWI, AWEIsh and WI2015 all use the NIR band in their calculations and had similar importance scores to this band.

\psection{Limitations and Future Work}

We interpreted a semantic segmentation model used to classify the pixels of coastline images into land and water. We used a permutation importance approach to understand which spectral bands were most important when predicting segmentation masks. This analysis showed that the NIR was the most important. Permuting this band lead to a decrease in accuracy 14 times higher than any other band. This is followed by Water Vapour, SWIR 1 and Blue bands which decreased the accuracy by 2.58,  0.78 and 0.19 percentage points respectively. 

The Water Vapour score was particularly interesting. The analysis suggests that this band is useful for extracting coastal water bodies. However, it is not typical for this band to be used in water extraction indices. Future research could investigate this potentially leading to a water extraction index using this band. 

The other spectral bands, Coastal Aerosol, Green, Red, RE1, RE2, RE3, RE4 and SWIR 2, were not important to the prediction. They produced a small increase in accuracy ($<0.5$ percentage points) both when permuted individually and when they when all permuted together. This suggests these bands could be removed from future model development without having a negative impact on model accuracy. At the same time, fewer data can reduce the complexity and computational requirements of future models. 

Lastly, we plan to explore other methods of interpreting semantic segmentation models. In this paper, we focused on the overall importance of spectral bands. Other methods, such as SHAP values, can be used to understand which pixels are important to a prediction. This could lead to new insights such as why the model is incorrectly predicting certain pixels.

\ack
This publication has emanated from research conducted with the financial support of Science Foundation Ireland under Grant number 18/CRT/6183. For the purpose of Open Access, the author has applied a CC BY public copyright licence to any Author Accepted Manuscript version arising from this submission. This research was conducted with the financial support of Science Foundation Ireland under Grant Agreement No.\ 13/RC/2106\_P2 at the ADAPT SFI Research Centre at University College Dublin. The ADAPT Centre for Digital Content Technology is partially supported by the SFI Research Centres Programme (Grant 13/RC/2106\_P2) and is co-funded under the European Regional Development Fund.

\bibliographystyle{IEEEbib}
\bibliography{./ref/longforms,./ref/references}

\end{paper}

\appendix

\section{Appendix}

\subsection{Remove images}
The images listed below were removed from the SWED test set for this analysis. The first image  was removed as the mask is flipped---land is labelled as 1 and water 0. The other two are removed as part of the land has not been labelled as land. 

\text{S2A\_MSIL2A\_20190803T025551\_N0213\_R032\_T54XWG\_20190803T043943\_image\_0\_0.tif}

\text{S2A\_MSIL2A\_20190901T101031\_N0213\_R022\_T34VDM\_20190901T130348\_image\_0\_0.tif}

\text{S2A\_MSIL2A\_20200405T100021\_N0214\_R122\_T34VDM\_20200405T115512\_image\_0\_0.tif}

\end{document}

%% file: figures/indices.tex
\begin{table}[h]
\centering
\begin{tabular}{|l|l|l|}
\hline
\multicolumn{1}{|c|}{\textbf{Indices}}                                                                                                                   & \multicolumn{1}{c|}{\textbf{Acronym}} & \multicolumn{1}{c|}{\textbf{\begin{tabular}[c]{@{}c@{}}Spectral bands \\ used in formula\end{tabular}}} \\ \hline
\begin{tabular}[c]{@{}l@{}}Normalized Difference \\ Water Index~\cite{mcfeeters1996use}\end{tabular}                               & NDWI                                  & NIR and Green                                                                                           \\ \hline
\begin{tabular}[c]{@{}l@{}}Automated Water \\ Extraction Index with \\ Shadows Elimination~\cite{feyisa2014automated}\end{tabular} & AWEIsh                                & \begin{tabular}[c]{@{}l@{}}Blue, Green, \\ NIR, SWIR1 \\ and SWIR2\end{tabular}                         \\ \hline
Water Index 2015~\cite{fisher2016comparing}                                                                                        & WI2015                                & \begin{tabular}[c]{@{}l@{}}Green, Red, \\ NIR, SWIR1 \\ and SWIR2\end{tabular}                          \\ \hline
Water Index 2~\cite{viana2019automatic}                                                                                            & WI2                                   & Blue and SWIR2                                                                                          \\ \hline
Sentinel-2 Water Index~\cite{jiang2021effective}                                                                                   & SWI                                  & RE1 and SWIR2                                                                                           \\ \hline
\end{tabular}
\caption{Indices used for water extraction. The last column gives the spectral bands used in the calculation for the given index. }
\label{tab:indices}
\end{table}

%% file: figures/confusion_matrix.tex
\begin{table}[h]
\centering
\begin{tabular}{lllll}
                                     &                                 & \multicolumn{2}{c}{\textbf{Prediction}}                                                     &                                          \\ \cline{3-4}
                                     & \multicolumn{1}{l|}{}           & \multicolumn{1}{l|}{\textbf{1}}              & \multicolumn{1}{l|}{\textbf{0}}              &                                          \\ \cline{2-5} 
\multicolumn{1}{c|}{\textbf{Actual}} & \multicolumn{1}{l|}{\textbf{1}} & \multicolumn{1}{l|}{True Positive (TP)}      & \multicolumn{1}{l|}{False Negative (FN)}     & \multicolumn{1}{l|}{Actual Positive (P)} \\ \cline{2-5} 
\multicolumn{1}{c|}{\textbf{Value}}  & \multicolumn{1}{l|}{\textbf{0}} & \multicolumn{1}{l|}{False Positive (FP)}     & \multicolumn{1}{l|}{True Negative (TN)}      & \multicolumn{1}{l|}{Actual Negative (N)} \\ \cline{2-5} 
\multicolumn{2}{l|}{}                                                  & \multicolumn{1}{l|}{Predicted Positive (P')} & \multicolumn{1}{l|}{Predicted Negative (N')} & \multicolumn{1}{l|}{Total (T)}           \\ \cline{3-5} 
\end{tabular}
\caption{Confusion matrix for pixel classification. Water pixels are represented by a value of 1 and land pixels are represented by a value of 0. }
\label{tab:confusion_matrix}
\end{table}

%% file: figures/eval_results.tex
\begin{table}[h]
\centering
\begin{tabular}{|l|l|l|l|l|l|}
\hline
\multicolumn{1}{|c|}{\textbf{Model}} & \multicolumn{1}{c|}{\textbf{Accuracy}} & \multicolumn{1}{c|}{\textbf{\begin{tabular}[c]{@{}c@{}}Balanced \\ Accuracy\end{tabular}}} & \multicolumn{1}{c|}{\textbf{Precision}} & \multicolumn{1}{c|}{\textbf{Recall}} & \multicolumn{1}{c|}{\textbf{F1}} \\ \hline
Original~\cite{seale2022swed}             & {\color[HTML]{2E2E2E} 0.937}           & {\color[HTML]{2E2E2E} 0.910}                                                               & {\color[HTML]{2E2E2E} 0.916}            & {\color[HTML]{2E2E2E} 0.948}         & {\color[HTML]{2E2E2E} 0.922}     \\ \hline
Replicated                           & 0.938                                  & 0.916                                                                                      & 0.924                                   & 0.954                               & 0.930                            \\ \hline
\end{tabular}
\caption{Evaluation metrics for the original and replicated model. The difference can likely be attributed to aspects of the model development process that could not be replicated exactly.}
\label{tab:eval}
\end{table}
